\documentclass[12pt, a4paper]{article}

\usepackage[margin=1in]{geometry}

\usepackage[T1]{fontenc}
\usepackage{graphicx}
\usepackage{array}
\usepackage{tabularray}
\usepackage{booktabs}
\usepackage{xcolor}
\usepackage{authblk} 
\usepackage{fontawesome5}
\usepackage{arydshln}

\newcolumntype{P}[1]{>{\raggedright\arraybackslash}p{#1}}
\newcolumntype{C}[1]{>{\centering\arraybackslash}p{#1}}

\usepackage{hyperref}
\hypersetup{
    colorlinks,
    linkcolor={red!50!black},
    citecolor={blue!50!black},
    urlcolor={green!50!black}
}

\begin{document}

\title{Benchmarking the Robustness of Foundation Models for Mammography under Domain Shift}

\author[1]{Giang Nguyen\textsuperscript{*, \faEnvelope[regular], }}
\author[2]{Raghav Mehta\textsuperscript{*, \faEnvelope[regular], }}
\author[2]{Emma A.M. Stanley}
\author[2]{Tian Xia}
\author[3]{Thi Hao Nguyen}
\author[1,4,5]{Hieu Pham}
\author[2]{Ben Glocker}

\affil[1]{\small College of Engineering and Computer Science, VinUniversity, Hanoi, Vietnam}
\affil[2]{\small Imperial College London, London, UK}
\affil[3]{\small Radiology Department, Vietnam National Cancer Hospital, Hanoi, Vietnam}
\affil[4]{\small VinUni-Illinois Smart Health Center, VinUniversity, Hanoi, Vietnam}
\affil[5]{\small The Computer Vision and Medical AI Lab, VinUniversity, Hanoi, Vietnam}

\affil[ ]{}
\affil[ ]{\small \textsuperscript{*}Equal contribution} 
\affil[ ]{}
\affil[ ]{\small \textsuperscript{\faEnvelope[regular]} Corresponding authors: 23giang.ns@vinuni.edu.vn, raghav.mehta@imperial.ac.uk}

\date{} 

\maketitle

\begin{abstract}
Foundation models are increasingly used as image feature extractors for mammography, but their robustness under external domain shift remains unclear. We benchmark 15 foundation-model backbones across breast density, BI-RADS severity, and cancer status using a unified frozen-backbone linear-probe protocol, training on 3 source datasets and evaluating on 12 task-compatible out-of-distribution (OOD) datasets after label harmonization. Mammography-specific vision-language models (\texttt{Mammo-FM} and \texttt{MaMA}) provide the strongest mean OOD performance, but robustness is not explained by mammography exposure alone. \texttt{DINOv3} remains a competitive vision-only baseline, and mammography-adapted pretraining does not consistently improve generalization. Dataset-level analysis further shows that even leading models show heterogeneous performance across datasets. Feature-space inspection reveals that useful representations can preserve clinical signal while retaining dataset and acquisition structure. These findings highlight dataset-level OOD evaluation as a central criterion for assessing mammography representations. Our code is publicly available at: 
{\footnotesize\url{https://github.com/biomedia-mira/mammo-ood}}.

\vspace{0.5cm}

\end{abstract}

\section{Introduction}
Foundation models (FMs) offer a practical route towards broadly applicable mammography representations. A typical workflow freezes a pretrained backbone, trains a lightweight classification head on a labeled source dataset, and applies the classifier to external data. A single backbone may support multiple downstream mammography tasks while reducing task-specific architectural design. However, the utility of an FM depends on whether its pretrained representation remains useful under distribution shift.

Prior work~\cite{bommasani2021,germani2025,glocker2023,moor2023,paschali2025} has shown that large-scale or modality-specific pretraining does not by itself guarantee robust out-of-distribution (OOD) generalization. This issue is amplified in mammography by heterogeneous clinical labels: density, BI-RADS, and cancer status differ in granularity, prevalence, and annotation protocols. Thus, in-distribution (ID) performance may not translate to OOD generalization.
To evaluate whether mammography representations generalize beyond their source data, we establish a benchmark for FM encoders across heterogeneous clinical tasks. We compare natural-image self-supervised learning (SSL)~\cite{dinov2,dinov3}, general radiology SSL~\cite{raydino,raddino}, mammography adapted SSL~\cite{mae,dinov2,dinov3,versamammo}, general medical vision-language models (VLM)~\cite{unimedclip,medsiglip,biomedclip}, and mammography specific VLM~\cite{mama,glam,mammoclip,mammofm} under frozen-backbone linear-probe protocol. Our contributions are threefold:

\begin{itemize}
\item We construct a harmonized benchmark from 15 public mammography datasets spanning 12 countries/regions. It covers three different downstream tasks: image-level density, exam-level BI-RADS, and exam-level cancer status.
\item We evaluate 15 foundation-model backbones under a unified frozen-backbone source-to-external ID/OOD protocol across these three tasks.
\item We analyze OOD behavior beyond aggregate scores by combining model-family comparisons, per-dataset evaluation, and feature-space inspection.
\end{itemize}

\begin{figure}[t]
\centering
\includegraphics[width=\textwidth]{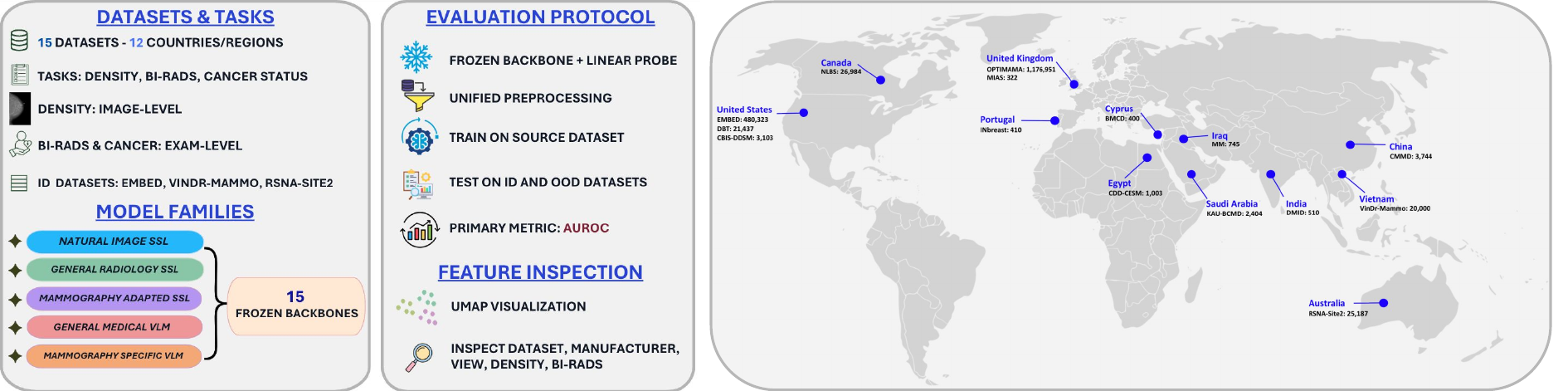}
\caption{Overview of the mammography foundation-model benchmark.}
\label{fig:benchmark_overview}
\end{figure}

Our results show that mammography pretraining alone is insufficient for robust OOD generalization. Instead, OOD generalization depends on the pretraining objective, model type (vision-only vs. vision-language), and pretraining dataset diversity. Mammography-specific vision-language models (VLMs) -- \texttt{MaMA}~\cite{mama} and \texttt{Mammo-FM}~\cite{mammofm} -- achieve the strongest overall performance, \texttt{DINOv3}~\cite{dinov3} remains a competitive vision-only baseline, and continued pretraining on a single mammography dataset does not improve OOD generalization. Together, these findings highlight the importance of OOD evaluation as a central criterion for assessing the clinical utility of FM representations in mammography.

\section{Benchmark Setup}

\subsection{Benchmark Datasets and Tasks}

We study three clinically distinct mammography prediction tasks: breast-density classification, BI-RADS assessment prediction~\cite{birads}, and cancer-status classification. These tasks differ in their clinical meaning and label structure: breast density characterizes breast tissue composition, BI-RADS provides an ordinal radiological assessment, and cancer status is a clinically important binary outcome with substantial class imbalance. As public datasets differ in their native annotation schemes, we harmonize labels before evaluation. BI-RADS and cancer status are evaluated per exam, reflecting clinical reading workflow, while breast density is evaluated per image.
  
The benchmark covers 15 datasets from 12 countries across North America, Europe, Asia, the Middle East, Africa, and Oceania. For each task, the source (ID) dataset acts as the training data for the task-specific linear classifier on frozen backbone features: \texttt{VinDr-Mammo} for BI-RADS, \texttt{EMBED} for density, and \texttt{RSNA-site2}\footnote{We only use \texttt{RSNA-site2} (the Australian cohort) and exclude \texttt{RSNA-site1} as it overlaps with \texttt{EMBED}, avoiding leakage across datasets.} for cancer status\footnote{Since \texttt{EMBED}, \texttt{VinDr}, and \texttt{RSNA} were used as unsupervised pretraining datasets for some foundational models (see Sec.~\ref{Sec.FM}), using them for OOD evaluation may introduce a small but non-negligible effect.}. As each dataset has different available labels, only task-compatible samples are included after filtering in Table~\ref{tab:dataset_counts}.   

\begin{table}[t]
\caption{Benchmark datasets after task-specific label filtering. Values are task-specific sample counts: exams for BI-RADS/cancer and images for density. * marks the source dataset for each task; parentheses show held-out ID test units for source datasets. \textdagger{} marks BI-RADS targets with available ordinal labels but a missing class; these targets are excluded from the final OOD AUROC result to avoid biasing the benchmark.}
\label{tab:dataset_counts}
\centering
\fontsize{8.8}{8.8}\selectfont
\setlength{\tabcolsep}{0pt}
\begin{tabular*}{\textwidth}{@{\extracolsep{\fill}}lccccccc@{}}
\toprule
 & \begin{tabular}{@{}c@{}}\scriptsize{\texttt{VinDr-Mammo}}\cite{vindr}\end{tabular}
 & \begin{tabular}{@{}c@{}}\scriptsize{\texttt{EMBED}}\cite{embed}\end{tabular}
 & \begin{tabular}{@{}c@{}}\scriptsize{\texttt{RSNA-site2}}\cite{rsna}\end{tabular}
 & \begin{tabular}{@{}c@{}}\scriptsize{\texttt{OPTIMAM}}\cite{optimam}\end{tabular}
 & \begin{tabular}{@{}c@{}}\scriptsize{\texttt{CBIS-DDSM}}\cite{cbis}\end{tabular}
 & \begin{tabular}{@{}c@{}}\scriptsize{\texttt{CDD-CESM}}\cite{cddcesm}\end{tabular}
 & \begin{tabular}{@{}c@{}}\scriptsize{\texttt{INbreast}}\cite{inbreast}\end{tabular} \\
\midrule
Region & Vietnam & USA & Australia & UK & USA & Egypt & Portugal \\ \midrule
BI-RADS 1 & *2,515 (494) & 8,897 & -- & -- & 1 & 47 & 10 \\
BI-RADS 2 & *1,568 (319) & 2,064 & -- & -- & 101 & 41 & 44 \\
BI-RADS 3 & *436 (91) & 723 & -- & -- & 185 & 60 & 9 \\
BI-RADS 4 & *368 (73) & 227 & -- & -- & 850 & 103 & 20 \\
BI-RADS 5 & *113 (23) & 33 & -- & -- & 309 & 73 & 25 \\ \midrule
Density A & 20 & *31,320 (6,674) & -- & 9,448 & 425 & 8 & 136 \\
Density B & 380 & *115,182 (22,565) & -- & 46,492 & 1,207 & 328 & 146 \\
Density C & 3,060 & *107,243 (22,545) & -- & 26,580 & 952 & 514 & 98 \\
Density D & 540 & *14,062 (2,727) & -- & 11,100 & 517 & 70 & 28 \\  \midrule
Non-cancer & 4,852 & 648 & *5,861 (1,156) & 46,569 & 814 & 147 & -- \\
Cancer & 113 & 298 & *234 (44) & 1,284 & 752 & 179 & -- \\
\bottomrule
\end{tabular*}

\begin{tabular*}{\textwidth}{@{\extracolsep{\fill}}lcccccccc@{}}
\toprule
 & \begin{tabular}{@{}c@{}}\scriptsize{\texttt{DMID}}\cite{dmid}\end{tabular}
 & \begin{tabular}{@{}c@{}}\scriptsize{\texttt{BMCD}}\cite{bmcd}\end{tabular}
 & \begin{tabular}{@{}c@{}}\scriptsize{\texttt{KAU-BCMD}}\cite{kau}\end{tabular}
 & \begin{tabular}{@{}c@{}}\scriptsize{\texttt{CMMD}}\cite{cmmd}\end{tabular}
 & \begin{tabular}{@{}c@{}}\scriptsize{\texttt{DBT}}\cite{dbt}\end{tabular}
 & \begin{tabular}{@{}c@{}}\scriptsize{\texttt{MIAS}}\cite{mias}\end{tabular}
 & \begin{tabular}{@{}c@{}}\scriptsize{\texttt{MM}}\cite{mm}\end{tabular}
 & \begin{tabular}{@{}c@{}}\scriptsize{\texttt{NLBS}}\cite{nlbs}\end{tabular} \\
\midrule
Region & India & Cyprus & Saudi Arabia & China & USA & UK & Iraq & Canada \\ \midrule
BI-RADS 1 & 209 & \textdagger{}22 & \textdagger{}460 & -- & -- & -- & -- & -- \\
BI-RADS 2 & 25 & \textdagger{}28 & \textdagger{}0 & -- & -- & -- & -- & -- \\
BI-RADS 3 & 121 & \textdagger{}0 & \textdagger{}102 & -- & -- & -- & -- & -- \\
BI-RADS 4 & 135 & \textdagger{}48 & \textdagger{}32 & -- & -- & -- & -- & -- \\
BI-RADS 5 & 19 & \textdagger{}2 & \textdagger{}9 & -- & -- & -- & -- & -- \\ \midrule
Density A & 79 & 20 & 680 & -- & -- & -- & -- & -- \\
Density B & 203 & 114 & 1,053 & -- & -- & -- & -- & -- \\
Density C & 186 & 56 & 383 & -- & -- & -- & -- & -- \\
Density D & 40 & 10 & 126 & -- & -- & -- & -- & -- \\ \midrule
Non-cancer & 180 & -- & -- & 465 & 5,389 & 270 & 620 & 5,848 \\
Cancer & 130 & -- & -- & 1,310 & 88 & 52 & 125 & 149 \\
\bottomrule
\end{tabular*}
\end{table}

\subsection{Foundation Models}\label{Sec.FM}

We evaluate 15 foundation-model backbones grouped into four non-overlapping families (Table~\ref{tab:model_families}). The mammography-adapted family consists of three vision-only SSL models that we further pretrained on the \texttt{EMBED}~\cite{embed} dataset. All other models are publicly available, which we directly used for feature extraction without any further modification. For \texttt{VersaMammo}, we evaluate the stage-1 backbone pretrained using the \texttt{DINOv2} framework, rather than its later distilled variant supervised by a CNN backbone. Reported mammography-specific training sources for various publicly available models are: \texttt{EMBED} for \texttt{MaMA} and \texttt{GLAM}; \texttt{UMPC} (private dataset) and \texttt{VinDr-Mammo} for \texttt{Mammo-CLIP}; \texttt{EMBED}, \texttt{UMPC} (private dataset), and \texttt{Mayo Clinic} (private dataset) for \texttt{Mammo-FM}; and \texttt{EMBED}, \texttt{VinDr-Mammo}, and \texttt{RSNA} for \texttt{VersaMammo}.

\begin{table}[!t]
\caption{Foundation-model families used throughout the paper. Family refers to pretraining/domain source; $^\dagger$ denotes VLMs.}
\label{tab:model_families}
\centering
\fontsize{9.8}{9.8}\selectfont
\setlength{\tabcolsep}{1pt}
\begin{tabular*}{\textwidth}{@{\extracolsep{\fill}}ll@{}}
    
\toprule
Family & Models \\
\midrule
Natural image SSL & \texttt{DINOv2}~\cite{dinov2}, \texttt{DINOv3}~\cite{dinov3} \\ \midrule

General radiology SSL & \texttt{RAD-DINO}~\cite{raddino}, \texttt{RayDINO}~\cite{raydino} \\ \midrule

Mammography adapted SSL & \texttt{DINOv2-EMBED}~\cite{dinov2}, \texttt{DINOv3-EMBED}~\cite{dinov3}, \texttt{MAE-EMBED}~\cite{mae}, \texttt{VersaMammo}~\cite{versamammo} \\ \midrule

General medical VLM & \texttt{BioMedCLIP}$^\dagger$~\cite{biomedclip}, \texttt{UniMedCLIP}$^\dagger$~\cite{unimedclip}, \texttt{MedSigLIP}$^\dagger$~\cite{medsiglip} \\ \midrule

Mammography specific VLM &\texttt{MaMA}$^\dagger$~\cite{mama}, \texttt{GLAM}$^\dagger$~\cite{glam}, \texttt{Mammo-CLIP}$^\dagger$~\cite{mammoclip}, \texttt{Mammo-FM}$^\dagger$~\cite{mammofm} \\

\bottomrule
\end{tabular*}
\end{table}

\subsection{Evaluation Protocol}
\label{eval}

All models are evaluated as frozen image encoders. For ViT-style models, we use the final classification-token or pooled representation as the frozen image feature; for CNN-based models, we use global-average-pooled features. A task-specific linear classifier is trained on the source (ID) dataset\footnote{For a fixed random seed, models were trained using the AdamW optimizer, a learning rate of 0.001, and binary cross-entropy loss for 20 epochs. For cancer classification, we used a weighted batch sampler with $\alpha = 0.3$ to tackle the high class imbalance.} and evaluated on the held-out ID test set and compatible external OOD datasets. All datasets are standardized through a common preprocessing pipeline, and images are resized to  $1024 \times 768$, preserving the global breast context. For BI-RADS and cancer status, image-level model probabilities are aggregated to the exam level by per-class max pooling followed by renormalization. Similarly, exam-level \textit{ground-truth} labels are defined by the maximum image-level BI-RADS or cancer label within each exam. Density predictions are evaluated directly at the image level. We use macro-averaged one-vs-rest AUROC for BI-RADS and density, and binary AUROC for cancer. Unless stated otherwise, OOD scores are averaged over task-compatible external datasets with finite AUROC. We calculate 95\% confidence intervals (CIs) using 1000 bootstrap runs (with replacement) for each dataset/model; $\pm$ denotes CI half-width.

\section{Results}

\begin{figure}[t]
\centering
\includegraphics[width=\textwidth]{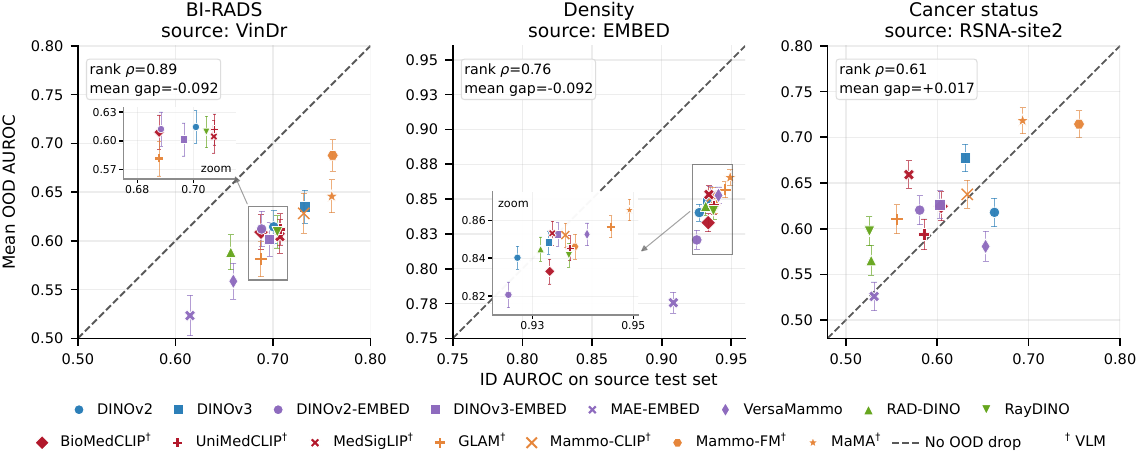}
\caption{ID-to-OOD generalization of foundation-model representations. Each point is one frozen backbone. Axes show ID AUROC (x-axis) and mean OOD AUROC (y-axis) across compatible external datasets. Error bars show 95\% confidence interval across OOD datasets. Colors indicate pretraining/domain family, and $^\dagger$ marks VLM encoders. The dashed diagonal marks equivalent ID and OOD performance.}
\label{fig:id_ood_scatter}
\end{figure}

\subsection{ID--OOD Performance Alignment and Gaps}

Fig.~\ref{fig:id_ood_scatter} plots ID AUROC against mean OOD AUROC for density, BI-RADS, and cancer status; the diagonal indicates equivalent ID and OOD performance. For BI-RADS, source and external performance are strongly rank-aligned (Spearman $\rho=0.89$), but models still fall below the diagonal on average, with a mean OOD gap of $-0.092$. \texttt{Mammo-FM} and \texttt{MaMA} occupy the high-ID/high-OOD region, while \texttt{DINOv3} is the strongest vision-only model. Density shows a higher source-domain ceiling, with many models reaching ID AUROC between 0.93 and 0.95. However, the mean OOD gap remains the same ($- 0.092$) with a lower rank correlation ($\rho = 0.76$), reflecting a narrow, high-performing cluster led by \texttt{MaMA}. Cancer has the weakest ID/OOD rank relationship ($\rho=0.61$) and a slightly positive mean gap (+0.017). These results are noisy because cancer-status targets differ in label definition across datasets: from BI-RADS-derived labels in \texttt{VinDr-Mammo} (1--2 neg., 5 pos.) to coarse normal/benign/abnormal annotations, and screening vs. cancer-enriched cohorts. Hence, the averaged score mixes results with heterogeneous task difficulty.

\subsection{Model-Level OOD Performance}

We next focus on external performance and ask which models remain strong after a domain shift (see Table~\ref{tab:ood_summary}).  Mammography-specific VLMs achieve the highest performance, with \texttt{MaMA}, \texttt{Mammo-FM}, and \texttt{Mammo-CLIP} among the strongest models. The leading model is task-dependent rather than universal: \texttt{Mammo-FM} achieves the best BI-RADS OOD AUROC (0.688{\tiny$\pm$0.016}), whereas \texttt{MaMA} reaches the best performance for density (0.865{\tiny$\pm$0.006}) and cancer status (0.718{\tiny$\pm$0.014}). \texttt{DINOv3} is the most important general-purpose vision-only model. Despite no mammography specific pretraining, it reaches 0.635{\tiny$\pm$0.017} on BI-RADS, 0.848{\tiny$\pm$0.006} on density, and 0.677{\tiny$\pm$0.015} on cancer, outperforming several mammography adapted or mammography specific models. These results suggest two conclusions: mammography specific VLMs provide the strongest OOD performance overall, but a strong natural-image vision-only encoder remains a difficult baseline to beat.

\begin{table}[t]
\caption{Mean OOD AUROC by model and task. Values are mean $\pm$ 95\% CI across OOD datasets. Models are grouped by pretraining/domain family. \textbf{Bold} marks the best model per task; \underline{underline} marks second best. $^\dagger$ denotes VLMs.}
\label{tab:ood_summary}
\centering
\fontsize{10.3}{10.3}\selectfont
\setlength{\tabcolsep}{2.5pt}
\begin{tabular*}{\textwidth}{@{\extracolsep{\fill}}llccc@{}}
\toprule
Family & Model & BI-RADS & Density & Cancer \\
\midrule
Natural image SSL & \texttt{DINOv2}~\cite{dinov2}                   & 0.614 {\tiny$\pm$ 0.017}             & 0.840 {\tiny$\pm$ 0.006}             & 0.618 {\tiny$\pm$ 0.016} \\
                  & \texttt{DINOv3}~\cite{dinov3}                   & 0.635 {\tiny$\pm$ 0.017}             & 0.848 {\tiny$\pm$ 0.006}             & 0.677 {\tiny$\pm$ 0.015} \\ \cmidrule{2-5}
                  & \texttt{Avg.}                                   & 0.625 {\tiny$\pm$ 0.012}             & 0.844 {\tiny$\pm$ 0.004}             & 0.648 {\tiny$\pm$ 0.011} \\ \midrule

General           & \texttt{RAD-DINO}~\cite{raddino}                & 0.588 {\tiny$\pm$ 0.018}             & 0.845 {\tiny$\pm$ 0.006}             & 0.565 {\tiny$\pm$ 0.016} \\
radiology SSL     & \texttt{Ray DINO}~\cite{raydino}                & 0.609 {\tiny$\pm$ 0.017}             & 0.842 {\tiny$\pm$ 0.006}             & 0.597 {\tiny$\pm$ 0.016} \\ \cmidrule{2-5}
                  & \texttt{Avg.}                                   & 0.599 {\tiny$\pm$ 0.012}             & 0.843 {\tiny$\pm$ 0.005}             & 0.581 {\tiny$\pm$ 0.011} \\ \midrule
                  
Mammography       & \texttt{DINOv2-EMBED}~\cite{dinov2}             & 0.612 {\tiny$\pm$ 0.018}             & 0.821 {\tiny$\pm$ 0.007}             & 0.620 {\tiny$\pm$ 0.016} \\
adapted SSL       & \texttt{DINOv3-EMBED}~\cite{dinov3}             & 0.601 {\tiny$\pm$ 0.017}             & 0.853 {\tiny$\pm$ 0.006}             & 0.626 {\tiny$\pm$ 0.015} \\
                  & \texttt{MAE-EMBED}~\cite{mae}                   & 0.523 {\tiny$\pm$ 0.021}             & 0.776 {\tiny$\pm$ 0.007}             & 0.526 {\tiny$\pm$ 0.016} \\ 
                  & \texttt{VersaMammo}~\cite{versamammo}           & 0.559 {\tiny$\pm$ 0.018}             & 0.853 {\tiny$\pm$ 0.006}             & 0.581 {\tiny$\pm$ 0.016} \\ \cmidrule{2-5} 
                  & \texttt{Avg.}                                   & 0.574 {\tiny$\pm$ 0.009}             & 0.825 {\tiny$\pm$ 0.003}             & 0.588 {\tiny$\pm$ 0.008} \\ \midrule

General           & \texttt{BioMedCLIP}$^\dagger$~\cite{biomedclip} & 0.609 {\tiny$\pm$ 0.018}             & 0.833 {\tiny$\pm$ 0.007}             & 0.625 {\tiny$\pm$ 0.016} \\
medical VLM       & \texttt{UniMedCLIP}$^\dagger$~\cite{unimedclip} & 0.612 {\tiny$\pm$ 0.016}             & 0.845 {\tiny$\pm$ 0.007}             & 0.594 {\tiny$\pm$ 0.017} \\
                  & \texttt{MedSigLIP}$^\dagger$~\cite{medsiglip}   & 0.604 {\tiny$\pm$ 0.017}             & 0.853 {\tiny$\pm$ 0.006}             & 0.659 {\tiny$\pm$ 0.015} \\ \cmidrule{2-5}
                  & \texttt{Avg.}                                   & 0.608 {\tiny$\pm$ 0.010}             & 0.844 {\tiny$\pm$ 0.004}             & 0.626 {\tiny$\pm$ 0.009} \\ \midrule
       
Mammography       & \texttt{Mammo-CLIP}$^\dagger$~\cite{mammoclip}  & 0.628 {\tiny$\pm$ 0.020}             & 0.852 {\tiny$\pm$ 0.006}             & 0.638 {\tiny$\pm$ 0.015} \\
specific VLM      & \texttt{GLAM}$^\dagger$~\cite{glam}             & 0.581 {\tiny$\pm$ 0.018}             & \underline{0.857 {\tiny$\pm$ 0.006}} & 0.610 {\tiny$\pm$ 0.016} \\
                  & \texttt{MaMA}$^\dagger$~\cite{mama}             & \underline{0.646 {\tiny$\pm$ 0.017}} & \textbf{0.865 {\tiny$\pm$ 0.006}}    & \textbf{0.718 {\tiny$\pm$ 0.014}} \\
                  & \texttt{Mammo-FM}$^\dagger$~\cite{mammofm}      & \textbf{0.688 {\tiny$\pm$ 0.016}}    & 0.846 {\tiny$\pm$ 0.007}             & \underline{0.714 {\tiny$\pm$ 0.015}} \\ \cmidrule{2-5}
                  & \texttt{Avg.}                                   & 0.636 {\tiny$\pm$ 0.009}             & 0.855 {\tiny$\pm$ 0.003}             & 0.670 {\tiny$\pm$ 0.008} \\ \bottomrule

\end{tabular*}
\end{table}

\subsection{Pretraining Factors in OOD Generalization}

Across model families, OOD performance varies with clinical-language alignment, pretraining source, and pretraining objective (see Table~\ref{tab:ood_summary}). VLMs achieve higher mean OOD AUROCs than vision-only models for BI-RADS (0.624{\tiny$\pm$0.007} vs. 0.593{\tiny$\pm$0.006}), density (0.850{\tiny$\pm$0.002} vs. 0.835{\tiny$\pm$0.002}), and cancer status (0.651{\tiny$\pm$0.006} vs. 0.601{\tiny$\pm$0.006}). However, this advantage is not uniform: \texttt{DINOv3} remains competitive with \texttt{MaMA} on BI-RADS (0.635{\tiny$\pm$0.017} vs. 0.646{\tiny$\pm$0.017}) and closely trails the leading VLMs on cancer status (0.677{\tiny$\pm$0.015} vs. 0.718{\tiny$\pm$0.014} and 0.714{\tiny$\pm$0.015} for \texttt{MaMA} and \texttt{Mammo-FM}). Thus, although VLMs perform better on average, strong SSL vision-only models remain competitive for OOD generalization.

When grouped by broad pretraining domain (see Table~\ref{tab:ood_summary}), natural image SSL models achieve mean OOD AUROCs of 0.625{\tiny$\pm$0.012}, 0.844{\tiny$\pm$0.004}, and 0.648{\tiny$\pm$0.011} for BI-RADS, density, and cancer status, respectively. This compares to 0.599{\tiny$\pm$0.012}, 0.843{\tiny$\pm$0.005}, and 0.581{\tiny$\pm$0.011} for general radiology SSL models, and 0.574{\tiny$\pm$0.009}, 0.825{\tiny$\pm$0.003}, and 0.588{\tiny$\pm$0.008} for mammography-adapted SSL models. Backbone- and objective-level comparisons reveal a similar pattern. \texttt{DINOv3-EMBED} provides only a marginal gain in density prediction over \texttt{DINOv3} (0.853{\tiny$\pm$0.006} vs. 0.848{\tiny$\pm$0.006}), while decreasing performance on both BI-RADS (0.601{\tiny$\pm$0.017} vs. 0.635{\tiny$\pm$0.017}) and cancer status (0.626{\tiny$\pm$0.015} vs. 0.677{\tiny$\pm$0.015}). Furthermore, \texttt{VersaMammo} performs strongly only on density (0.853{\tiny$\pm$0.006}), despite its multi-source mammography pretraining. These results indicate that mammography exposure alone does not guarantee robust OOD generalization. 

On average, however, mammography-specific VLMs (0.636{\tiny$\pm$0.009}, 0.855{\tiny$\pm$0.003}, and 0.670{\tiny$\pm$0.008} for BI-RADS, density, and cancer status) outperform general medical VLMs (0.608{\tiny$\pm$0.010}, 0.844{\tiny$\pm$0.004}, and 0.626{\tiny$\pm$0.009} for BI-RADS, density, and cancer status). This demonstrates that mammography language alignment is more important than visual mammography exposure alone. Overall, these results indicate that OOD generalization depends on the interaction between the source domain, pretraining objective, and model type, rather than on mammography exposure alone.

\subsection{Dataset-Level OOD Heterogeneity}

\begin{figure}[t]
\centering
\includegraphics[width=\textwidth]{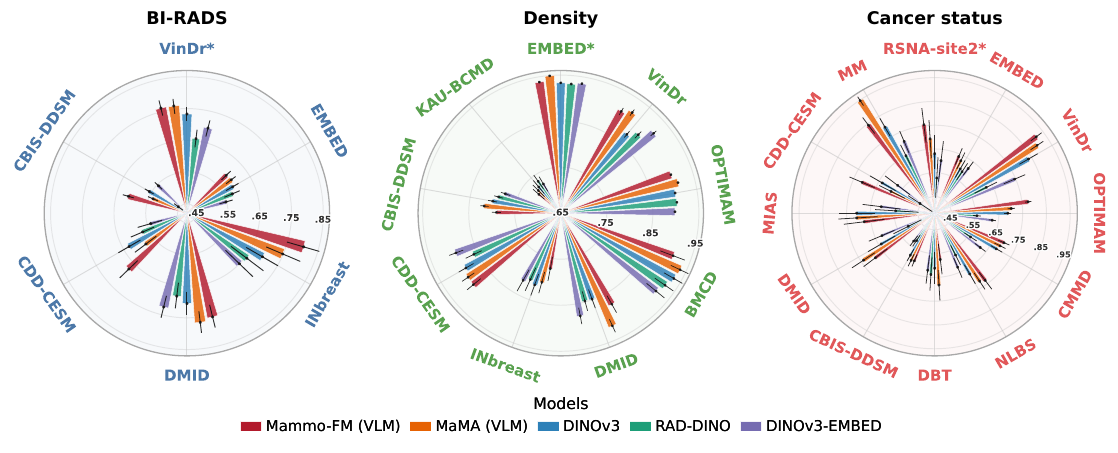}
\caption{Dataset-level AUROC performances (with 95\% confidence interval error-bar) for five representative backbones across BI-RADS, breast density, and cancer status. The ID datasets are marked with an asterisk ($\ast$); all other datasets are used as OOD. }
\label{fig:dataset_spider}
\end{figure}

To more closely assess how different backbones generalize across dataset shifts, we examine per-dataset performance for five representative models: \texttt{DINOv3}, \texttt{RAD-DINO}, \texttt{DINOv3-EMBED}, \texttt{MaMA}, and \texttt{Mammo-FM}. The first three provide a \texttt{DINO}-family contrast spanning natural-image, general-radiology, and mammography-adapted pretraining, while \texttt{MaMA} and \texttt{Mammo-FM} represent the strongest mammography specific VLMs. Fig.~\ref{fig:dataset_spider} shows that strong average OOD performance does not necessarily correspond to uniformly strong behavior across OOD datasets. All three \texttt{DINO}-family vision models exhibit heterogeneous performance across datasets and tasks. However, their relative ranking remains consistent, with \texttt{RAD-DINO} showing the lowest performance and \texttt{DINOv3} achieving the highest.

\begin{figure}[t]
\centering
\includegraphics[width=1\textwidth]{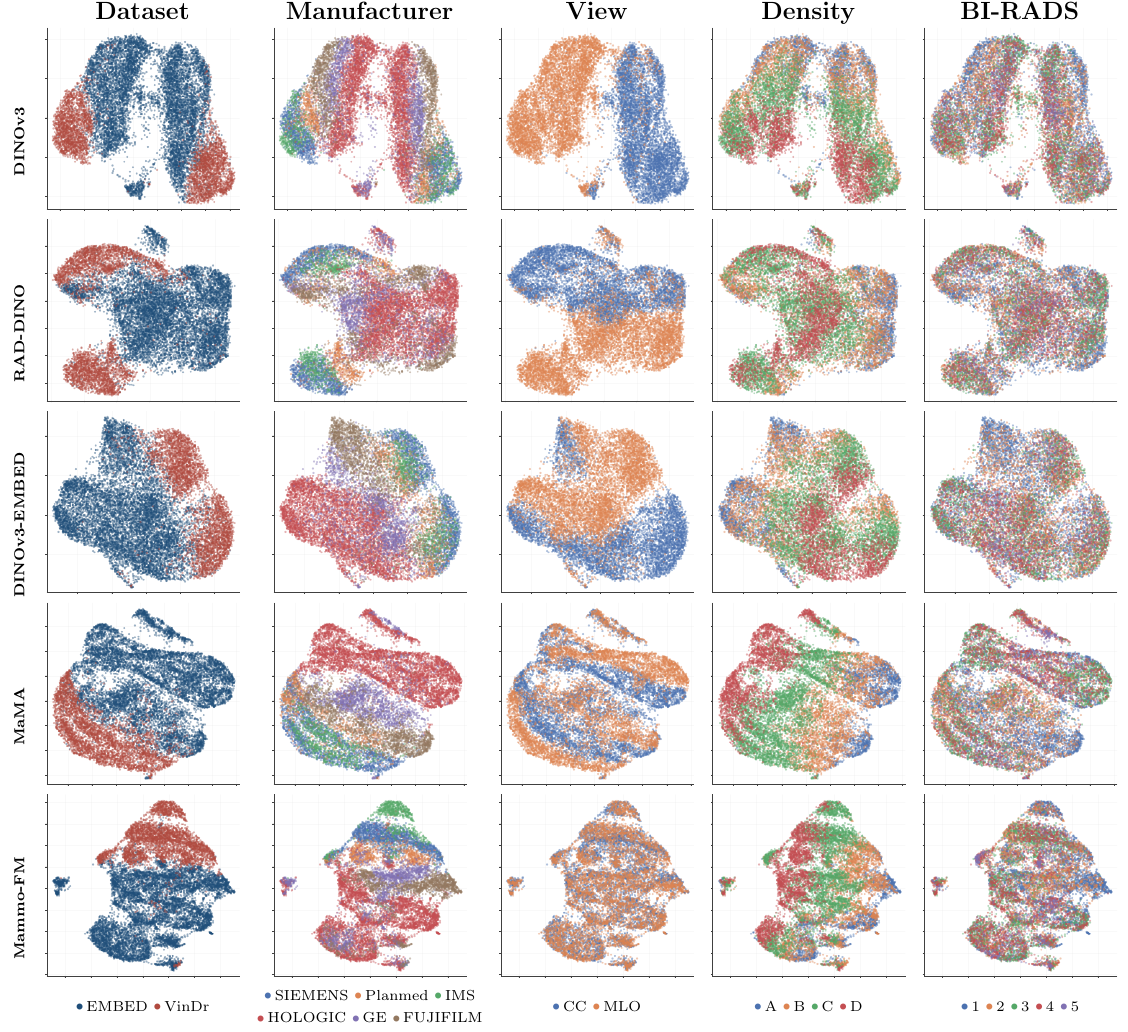}
\caption{UMAP inspection grid for five representative backbones. Rows are (top to bottom) \texttt{DINOv3}, \texttt{RAD-DINO}, \texttt{DINOv3-EMBED}, \texttt{MaMA}, and \texttt{Mammo-FM}. Columns are color-embedded by (left to right) dataset, manufacturer, view position, breast density, and BI-RADS.}
\label{fig:inspection_grid}
\end{figure}

For BI-RADS, \texttt{Mammo-FM} shows a higher mean OOD performance compared to \texttt{MaMA} (0.688{\tiny$\pm$0.016} vs. 0.646{\tiny$\pm$0.017}). However, Fig.~\ref{fig:dataset_spider} reveals that \texttt{Mammo-FM} demonstrates substantially higher performance on the \texttt{CDD-CESM} and \texttt{CBIS-DDSM} datasets, while performing similarly to \texttt{MaMA} on the other three datasets (\texttt{EMBED}, \texttt{DMID}, and \texttt{INbreast}). This performance disparity indicates \texttt{Mammo-FM}'s better overall generalization to variability in image acquisition parameters, as both \texttt{CDD-CESM} (contrast-enhanced mammography) and \texttt{CBIS-DDSM} (scanned film mammography) represent non-standard mammography modalities. However, for density, \texttt{MaMA} consistently outperforms \texttt{Mammo-FM} across all datasets (mean OOD 0.865{\tiny$\pm$0.006} vs 0.846{\tiny$\pm$0.007}). We hypothesise that this is due to the explicit use of density information during \texttt{MaMA}'s pretraining, as it relied on tabular metadata (e.g., density, BI-RADS) to generate text reports for language pretraining. In contrast, \texttt{Mammo-FM} used actual clinical reports, which do not explicitly encode density information, focusing instead on clinical outcomes like BI-RADS and cancer status. This difference in pretraining supervision is also reflected in cancer status. \texttt{Mammo-FM} performs better on datasets where cancer status is closer to an exam-level clinical outcome or proxy label (\texttt{RSNA}, \texttt{OPTIMAM}, \texttt{VinDr-Mammo}, and \texttt{CDD-CESM}), consistent with its report-based pretraining. Conversely, \texttt{MaMA} is stronger on datasets with more direct or curated cancer annotations (\texttt{DBT}, \texttt{DMID}, \texttt{MIAS}, and \texttt{MM}), suggesting that metadata-derived report supervision may better preserve explicit visual abnormality cues. 

\subsection{Clinical and Domain Structure in Feature Space}

Following the dataset-level analysis, we use UMAP~\cite{umap} as a qualitative inspection of what information remains organized in the same five representations. Fig.~\ref{fig:inspection_grid} compares their joint \texttt{EMBED}/\texttt{VinDr} embeddings. Across models, the dataset structure remains visible, indicating that strong external generalization does not require complete domain invariance. The \texttt{DINO}-family comparison shows that changing the pretraining source reshapes the embedding geometry but does not necessarily improve clinical organization. \texttt{DINOv3} preserves strong view structure and some density structure, while \texttt{RAD-DINO} and \texttt{DINOv3-EMBED} do not show clear density separation, consistent with their weaker OOD gains.

The two strongest mammography-specific VLMs show different qualitative biases. \texttt{MaMA} has the clearest density organization, consistent with its leading density OOD AUROC, and with report-aligned pretraining capturing density semantics. However, \texttt{MaMA} still preserves visible view separation. This was surprising, as \texttt{MaMA} explicitly aligned different views of the same patient during its image pre-training. We hypothesize that this is a byproduct of using tabular image information (e.g., view, side) in the language pre-training of the \texttt{MaMA} model. In contrast, \texttt{Mammo-FM} shows better alignment between different views, despite not having any explicit multi-view alignment in its pre-training. We hypothesize that this is due to the use of actual clinical reports during the language pre-training of \texttt{Mammo-FM}, which do not have image information encoded in them.

\section{Conclusion}

This benchmark shows that strong source-domain performance and mammography exposure alone do not guarantee robust OOD generalization. Mammography-specific VLMs achieve the strongest mean OOD performance, with \texttt{MaMA} and \texttt{Mammo-FM} forming the most consistent top pair, but their gains are task- and dataset-dependent. \texttt{DINOv3} remains a competitive vision-only baseline, while general-radiology pretraining and single-source mammography adaptation do not consistently improve transfer. These results suggest that robust mammography representations depend on how pretraining objectives and data sources align with each clinical task and its label provenance. Dataset-level and feature-space analyses further show that external robustness is heterogeneous rather than uniform. Therefore, mammography foundation models should be assessed using task-specific OOD validation rather than ID performance. 

A limitation of this study is that our linear-probe protocol measures representation quality rather than the best achievable absolute performance for different tasks. Future work should evaluate full fine-tuning and other model adaptation strategies to explore the full potential of different foundation models. In the future, it would also be interesting to evaluate uncertainty estimation and confidence calibration of foundational models across OOD datasets.

\section*{Acknowledgement}
This project was supported by the Foreign, Commonwealth \& Development Office (FCDO), Natural Sciences and Engineering Research Council (NSERC) of Canada, the Royal Academy of Engineering as part of the Kheiron/RAEng Research Chair, the Imperial College London UKRI Impact Acceleration Account \texttt{EP/X52556X/1}, and VinUniversity’s Seed Grant Program under Project \texttt{VUNI.2425.EME.005}.

\subsection*{Disclosure of interests}
B.G. is a part-time employee of DeepHealth. No other competing interests.


\end{document}